\pdfoutput=1
\documentclass[10pt, a4paper]{article}
\usepackage{lrec}
\usepackage{multibib}
\newcites{languageresource}{Language Resources}
\usepackage{graphicx}
\usepackage{tabularx}
\usepackage{soul}
\PassOptionsToPackage{hyphens}{url}
\usepackage{hyperref}
\usepackage{xstring}
\usepackage{color}

\usepackage[utf8]{inputenc}

\usepackage{url}
\usepackage{booktabs}
\usepackage{graphicx}   % for resizebox
\usepackage{subcaption}

\usepackage{amsmath}
\usepackage{amssymb}
\usepackage{ifthen}
\usepackage{booktabs}   % pretty tables
\usepackage{microtype}  % pretty typography
\usepackage{dsfont}     % for mathds
\usepackage[multidot]{grffile}  % dots in included pdf file names
\usepackage{rotating}

\DeclareMathOperator*{\argmax}{arg\,max}
\DeclareMathOperator*{\argmin}{arg\,min}

\newcommand{\prob}{P}

\newcommand{\vb}{\,|\,}

\newcommand{\params}{\boldsymbol{\theta}}
\newcommand{\data}{\boldsymbol{D}}

\newcommand{\bnd}{\mathbin{+\mkern-10mu+}}
\newcommand{\tbnd}{$\bnd$}

\newcommand{\hparams}{\hat{\params}}
\newcommand{\morph}{m}

\newcommand{\morphseq}{\boldsymbol{s}}
\newcommand{\fixme}[2][]{%
    \ifthenelse{\equal{#1}{}}%
               {\textsl{[\textcolor{red}{#2}]}}%
               {\textsl{[#1: \textcolor{red}{#2}]}}%
    }%
\newcommand{\term}[1]{\textbf{\emph{#1}}}
\newcommand{\examp}[2][]{\emph{``#2''}\ifthenelse{\equal{#1}{}}{}{ (#1)}}
\newcommand{\techterm}[1]{\textsc{#1}}
\newcommand{\abbr}[2][]{%
    \ifthenelse{\equal{#1}{}}%
               {#2}%
               {#1 (#2)}}%

\usepackage{numprint}
\npdecimalsign{.}
\npthousandsep{\ensuremath{\;}}
\npfourdigitnosep

\newcommand{\vh}[1]{\rotatebox{90}{#1}}

\newcommand{\bb}{\fontseries{b}\selectfont}
\newcommand{\sci}[1]{{\scriptsize $\times$10}$^{#1}$}

\newcommand{\uua}[0]{$\boldsymbol{\uparrow}\,$}
\newcommand{\dda}[0]{$\boldsymbol{\downarrow}\,$}

\newcommand{\nonsE}{$\sim$\textbf{E}}
\newcommand{\nonsB}{$\sim$\textbf{B}}

\newcommand{\subfigcost}[2]{
\begin{subfigure}[t]{0.49\textwidth} \centering
\includegraphics[width=#2\textwidth]{figures/print.#1.cost.loglog.x_y.low.png}
\end{subfigure}
}

\newcommand{\costsubfigs}{
\begin{figure*}[t]
\centering
\subfigcost{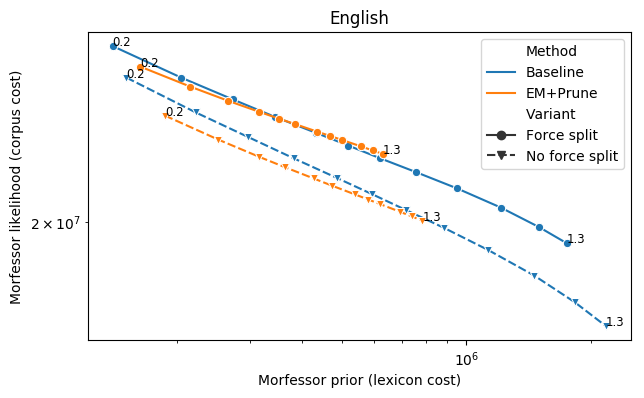}{1.0}
\subfigcost{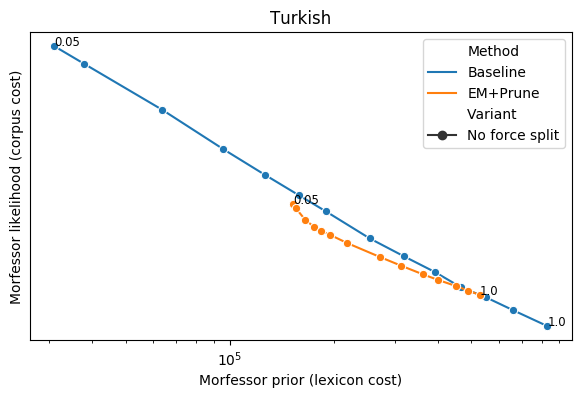}{0.9}
\\
\subfigcost{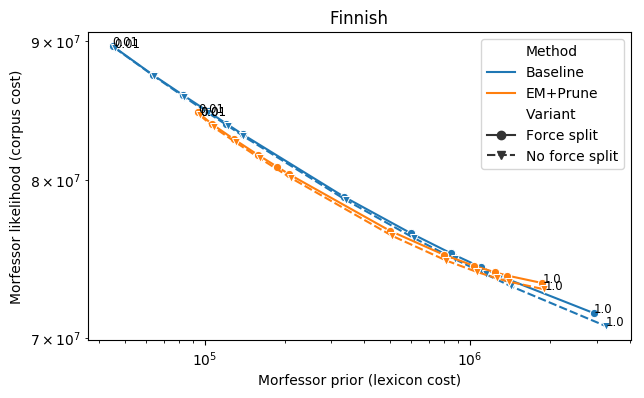}{1.0}
\subfigcost{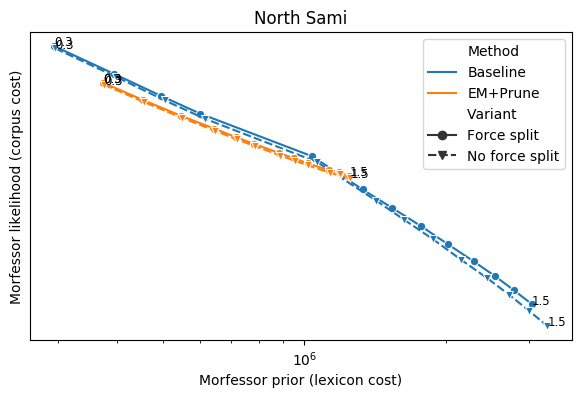}{0.9}
\caption{Unweighted Morfessor cost function components (prior and likelihood). Log scale.
\label{fig:costs}}
\end{figure*}
}

\newcommand{\subfigbpr}[2]{
\begin{subfigure}[t]{0.47\textwidth} \centering
\includegraphics[width=#2\textwidth]{figures/print.#1.bpr.x_y.png}
\end{subfigure}
}
\newcommand{\bprsubfigs}{
\begin{figure}[t]
\centering
\subfigbpr{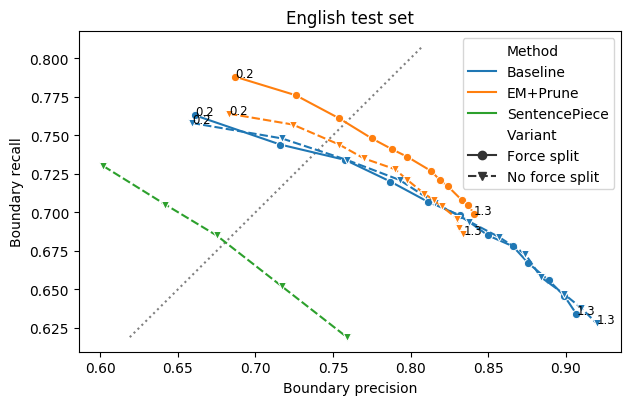}{0.98}
\subfigbpr{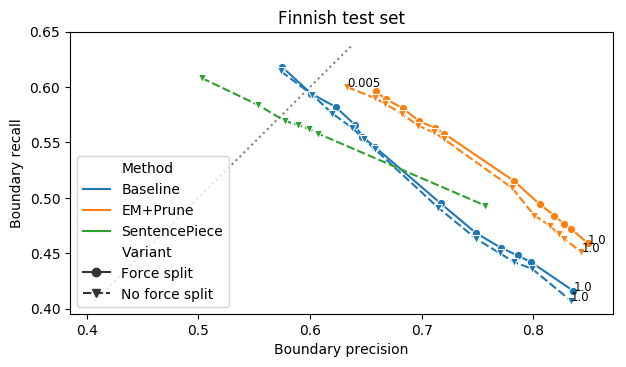}{0.98}
\subfigbpr{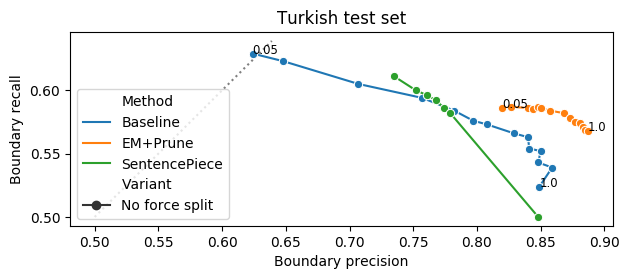}{0.98}
\subfigbpr{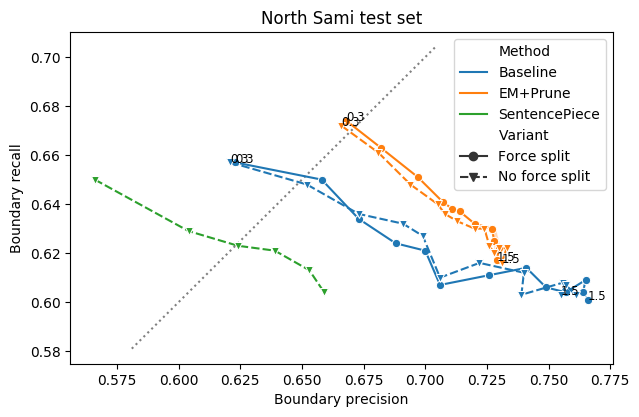}{0.98}
\caption{Boundary Precision--Recall curve at different tuning points,
The smallest and largest $\alpha$-values are labeled.
\label{fig:bprs}}
\end{figure}
}

\newcommand{\textnumero}{No}

\usepackage{enumitem}
\setenumerate{noitemsep,topsep=0.7ex,parsep=0.7ex,partopsep=0.7ex}

\title{Morfessor EM+Prune: Improved Subword Segmentation with Expectation Maximization and Pruning
}

\name{Stig-Arne Grönroos$^1$, Sami Virpioja$^2$, Mikko Kurimo$^1$}

\address{$^1$Department of Signal Processing and Acoustics, Aalto University, Finland \\
         $^2$Department of Digital Humanities, University of Helsinki, Finland \\
         \{stig-arne.gronroos,mikko.kurimo\}@aalto.fi, sami.virpioja@helsinki.fi \\}

\date{\today}

\abstract{Data-driven segmentation of words into subword units has
  been used in various natural language processing applications such
  as automatic speech recognition and statistical machine translation
  for almost 20 years. Recently it has became more widely adopted, as
  models based on deep neural networks often benefit from subword
  units even for morphologically simpler languages. In this paper, we
  discuss and compare training algorithms for a unigram subword model,
  based on the Expectation Maximization algorithm and lexicon
  pruning. Using English, Finnish, North Sami, and Turkish data sets,
  we show that this approach is able to find better solutions to the
  optimization problem defined by the Morfessor Baseline model than
  its original recursive training algorithm. The improved optimization
  also leads to higher morphological segmentation accuracy when
  compared to a linguistic gold standard. We publish implementations
  of the new algorithms in the widely-used Morfessor software
  package. \\ \newline
}
\begin{document}
% use \cite{} and \newcite{}

% terminology:
% cost function vs loss function?   Going to use *cost*, following prior Morfessor work.
% prior/likelihood vs lexicon/corpus cost?  Going to use *prior/likelihood*, following prior Morfessor work.
% vocabulary vs lexicon?    Going to use *lexicon*, following prior Morfessor work.
% morph vs subword?         Going to use both according to context.
% EM/Viterbi/Lateen-EM vs soft-EM/hard-EM/lateen-EM     Going to use a compromise: EM/hard-EM/lateen-EM
% Lateen-prune: Half-lateen? Viterbi-prune? Soft-EM Hard-Prune? Going to use EM+Viterbi-prune
% F-measure vs F-score

\maketitleabstract

\section{Introduction}
Subword segmentation has become a standard preprocessing step
in many neural approaches to natural language processing (NLP) tasks,
e.g Neural Machine Translation (NMT)~\cite{sennrich2015neural}
and Automatic Speech Recognition (ASR)~\cite{smit2017improved}.
Word level modeling suffers from sparse statistics,
issues with Out-of-Vocabulary (OOV) words,
and heavy computational cost due to a large vocabulary.
Word level modeling is particularly unsuitable for morphologically rich languages,
but subwords are commonly used for other languages as well.
Subword segmentation is best suited for languages with agglutinative morphology.

While rule-based morphological segmentation systems can achieve high quality,
the large amount of human effort needed makes the approach problematic,
particularly for low-resource languages.
The systems are language dependent, necessitating use of multiple tools in multilingual setups.
As a fast, cheap and effective alternative,
data-driven segmentation can be learned in a completely unsupervised manner from raw corpora. %unannotated corpora.

Unsupervised morphological segmentation saw much research interest until the early 2010's; for a survey on the methods, see \newcite{hammarstrom2011unsupervised}.
Semi-supervised segmentation with already small amounts of annotated training data was found to improve the accuracy significantly when compared to a linguistic segmentation; see \newcite{ruokolainen2016comparative} for a survey.
While this line of research has been continued in supervised and more grammatically oriented tasks \cite{cotterell-etal-2017-conll},
the more recent work on unsupervised segmentation is less focused on approximating a linguistically motivated segmentation.
Instead, the aim has been to tune subword segmentations for particular applications.
For example,
the simple substitution dictionary based
Byte Pair Encoding segmentation algorithm~\cite{gage1994}, first proposed for NMT by \newcite{sennrich2015neural}, has become a standard in the field.
Especially in the case of multilingual models,
training a single language-independent subword segmentation method is preferable to linguistic segmentation~\cite{arivazhagan2019massively}.

In this study, we compare three existing and one novel subword segmentation method,
all sharing the use of a unigram language model in a generative modeling framework.
The previously published methods are
Morfessor Baseline~\cite{creutz2002unsupervised},
Greedy Unigram Likelihood~\cite{varjokallio2013learning}, and
SentencePiece~\cite{kudo2018subword}.
The new Morfessor variant proposed in this work is called Morfessor EM+Prune.

% Like other Morfessor methods, Morfessor EM+Prune is applicable also
% to other segmentation tasks than segmenting words into morphs.
% We consider such tasks out of scope for this article.

The contributions of this article%
\footnote{
    This work is licensed under a Creative Commons Attribution–NoDerivatives
    4.0 International Licence.  Licence details:
    \scriptsize\url{http://creativecommons.org/licenses/by-nd/4.0/}}
are
\begin{enumerate}
\item[(i)] a better training algorithm for Morfessor Baseline,
with reduction of search error during training, % for all languages
and improved segmentation quality for English, Finnish and Turkish; % but not not North Sami
\item[(ii)] comparing four similar segmentation methods,
including a close look at the SentencePiece reference implementation,
highlighting details omitted from the original article \cite{kudo2018subword};
\item[(iii)] and showing that the proposed Morfessor EM+Prune with particular hyper-parameters yields SentencePiece.
\end{enumerate}

\begin{table*}[t]
\centering
\resizebox{0.88\textwidth}{!}{
\begin{tabular}{lllll}
\toprule
                                & \bb Morfessor BL  & \bb Greedy Unigram    & \bb SentencePiece   & \bb Morfessor EM+Prune  \\
\midrule
Model                           & Unigram LM        & Unigram LM            & Unigram LM          & Unigram LM          \\
\midrule
Cost function                   & MAP               & ML                    & MAP                 & MAP                 \\
\;  Prior                       & MDL               & --                    & DP                  & MDL+DP              \\
\midrule
Training algorithm              & Local search      & EM+Prune              & EM+Prune            & EM+Prune            \\
\;  Initialization              & Words             & Seed lexicon          & Seed lexicon        & Seed lexicon        \\
\;  EM variant                  & --                & Lateen-EM once        & EM                  & EM / Lateen-EM      \\
\midrule
Stopping criterion \\
\;  Cost change threshold       & \checkmark        & \checkmark            & --                  & \checkmark          \\
\;  Target lexicon size         & Approximate       & \checkmark            & \checkmark          & \checkmark          \\
\midrule
N-best decoding                 & \checkmark        & --                    & \checkmark          & \checkmark          \\
Sampling decoding               & --                & --                    & \checkmark          & \checkmark          \\
\midrule
Count dampening                 & \checkmark        & --                    & --                  & \checkmark          \\
Semi-supervised                 & \checkmark        & --                    & --                  & \checkmark          \\
Requires pretokenization        & \checkmark        & \checkmark            & --                  & \checkmark          \\
\midrule
Reference implementation        & Python            & C++                   & C++                 & Python              \\
\bottomrule

\end{tabular}}
\caption{Comparison of subword segmentation methods applying a unigram language model.
\label{tab:comparison}}
\end{table*}

\subsection{Morphological Segmentation with Unigram Language Models}

% The task of Morphological segmentation
Morphological surface segmentation is the task of splitting words
into morphs, the surface forms of meaning-bearing sub-word units, morphemes.
The concatenation of the morphs is the word, e.g.
\[ \quad reliability \mapsto reli \bnd abil \bnd ity \]
% probabilistic generative methods
Probabilistic generative methods for morphological segmentation
model the probability $\prob(\morphseq)$
of generating a sequence of morphs (a word, sentence or corpus)
$\morphseq = [\morph_{0}, \mathellipsis, \morph_{N}]$,
as opposed to discriminative methods
that model the conditional probability of the segmentation boundaries given the unsegmented data.
%$\prob(\boldsymbol{y} \vb \data)$.

% unigram LM
This study focuses on segmentation methods applying a \term{unigram language model}.
In the unigram language model,
an assumption is made that the morphs in a word occur independently of each other.
Alternatively stated, it is a zero-order (memoryless) Markov model,
generalized so that one observation can cover multiple characters.
The probability of a sequence of morphs
decomposes into the product of the probabilities of the morphs of which it consists.
\begin{equation}
\prob_{\params}(\morphseq) = \prod_{i=1}^{N} \prob_{\params}(\morph_{i})
\end{equation}

% Expectation Maximization
The Expectation Maximization (EM) algorithm~\cite{dempster1977maximum}
is an iterative algorithm for finding Maximum Likelihood (ML) or Maximum a Posteriori (MAP) estimates
for parameters in models with latent variables.
The EM algorithm consists of two steps.
In the E-step \eqref{eq:estep}, the expected value of the complete data likelihood including the latent variable is taken,
and in the M-step \eqref{eq:mstep}, the parameters are updated to maximize the expected value of the E-step:
\begin{align}
Q(\params, \params^{(i-1)}) &= \int_{\boldsymbol{y}} \log\prob(\data, \boldsymbol{y} \vb \params) \prob(\boldsymbol{y} \vb \data, \params^{(i-1)}) d\boldsymbol{y}
\label{eq:estep} \\
%& \txt{E-step} \\
\params^{i} &= \argmax_{\params} Q(\params, \params^{(i-1)})
\label{eq:mstep}.
%& \txt{M-step}
\end{align}

When applied to a (hidden) Markov model, EM is called the forward-backward algorithm.
Using instead the related Viterbi algorithm~\cite{viterbi1967error}
is sometimes referred to as \emph{hard-EM}.\footnote{An analogy can be drawn to clustering using $k$-means, which yields a hard assignment of data points to clusters, and using EM for clustering with a Gaussian Mixture Model (GMM), where the assignment is soft.}
\newcite{spitkovsky2011lateen} present lateen-EM,
a hybrid variant in which EM and Viterbi optimization are alternated.
%We use the term \emph{soft EM} to denote classical EM without the Lateen variant,
%and use \emph{EM} in an extended sense to include the variants.
%\cite{liang2007structured}

% Challenges of applying EM
\newcite[Section 6.4.1.3]{virpioja2012learning} discusses the challenges of applying
%Expectation Maximization (EM)
EM to learning of generative morphology.
Jointly optimizing both the morph lexicon and the parameters for the morphs is intractable.
If, like in Morfessor Baseline, the cost function is discontinuous
when morphs are added or removed from the lexicon, there is no closed form solution to the M-step.
%With Maximum Likelihood (ML)
With ML
estimates for morph probabilities,
EM can neither add nor remove morphs from the lexicon,
because it can neither change a zero probability to nonzero nor vice versa.
%If a morph has zero probability at any point,
%then any segmentation using that morph will have zero probability and
%the expected count of the morph will remain at zero.
%If a morph has a nonzero probability, the expected count will remain nonzero.

% local search
One solution to this challenge is to apply local search.
Starting from the current best estimate for the parameters,
small search steps are tried to explore near-lying parameter configurations.
The choice that yields the lowest cost is selected as the new parameters.
Greedy local search often gets stuck in local minima.
Even if there are parameters yielding a better cost, the search may not find them, causing search error.
The error remaining at the parameters with globally optimal cost is the model error.

% seed+pruning
Another solution is to combine EM with pruning (EM+Prune).
The methods based on pruning begin with a seed lexicon,
which is then iteratively pruned until a stopping condition is reached.
Subwords cannot be added to the lexicon after initialization.
As a consequence, proper initialization is important,
and the methods should not prune too aggressively without reestimating parameters,
as pruning decisions cannot be backtracked.
For this reason, EM+Prune methods proceed iteratively,
only pruning subwords up to a predefined iteration pruning quota,
e.g. removing at most 20\% of the remaining lexicon at a time.

\section{Related Work}
In this section we review three previously published segmentation methods
that apply a unigram language model.
Table~\ref{tab:comparison} summarizes the differences between these methods.
% present the following 4 sections all in the same order: cost, seed, training, other

\subsection{Morfessor Baseline}
Morfessor is a family of generative models for unsupervised morphology induction \cite{creutz07acmtslp}.
Here, consider the Morfessor 2.0 implementation~\cite{virpioja2013morfessor} of Morfessor Baseline method~\cite{creutz2002unsupervised}.

A point estimate for the model parameters $\hparams$ is found using
%Maximum A Posteriori
MAP estimation with a Minimum Description Length (MDL)~\cite{rissanen1989stochastic}
inspired prior that favors lexicons containing fewer, shorter morphs.
The MAP estimate yields a two-part cost function,
consisting of a prior (the lexicon cost) and likelihood (the corpus cost).
The model can be tuned using the hyper-parameter $\alpha$,
which is a weight applied to the likelihood~\cite{kohonen2010semi-supervised}:
\begin{equation}
%\hparams = \argmin_{\params} \cost(\params, \data) = \argmin_{\params} - \log\prob(\params) - \alpha\log\prob(\data \vb \params)
\hparams = \argmin_{\params} \{
    - \log\overbrace{\prob(\params) }^\text{prior} \;
    - \alpha\log\overbrace{\prob(\data \vb \params) }^\text{likelihood} \}
\label{eq:morfessorcost}
\end{equation}
The $\alpha$ parameter controls the overall amount of segmentation,
with higher values increasing the weight of each emitted morph in the corpus (leading to less segmentation),
and lower values giving a relatively larger weight to a small lexicon
%(leading to
(more segmentation). 

The prior can be further divided into two parts:
the prior for the morph form properties and the usage properties.
The form properties encode the string representation of the morphs,
while the usage properties encode their frequencies.
% takes too much room: full generative story of drawing the lexicon?
Morfessor Baseline applies a non-informative prior for the distribution of the morph frequencies.
It is derived using combinatorics
from the number of ways that the total token count $\nu$ can be divided among the $\mu$ lexicon items:
\begin{equation}
\prob(\tau_{1}, \mathellipsis, \tau_{\mu} \vb \mu, \nu) = 1 / \binom{\nu -1}{\mu - 1}.
\label{eq:freqdistr}
\end{equation}

Morfessor Baseline is initialized with a seed lexicon of whole words.
The Morfessor Baseline training algorithm is a greedy local search.
During training, in addition to storing the model parameters,
the current best segmentation for the corpus is stored in a graph structure.
The segmentation is iteratively refined,
by looping over all the words in the corpus in a random order and resegmenting them.
The resegmentation is applied by recursive binary splitting,
leading to changes in other words that share intermediary units with the word currently being resegmented.
The search converges to a local optimum,
and is known to be sensitive to the initialization~\cite{virpioja2013morfessor}.

In the Morfessor 2.0 implementation,
the likelihood weight hyper-parameter $\alpha$ is set either
with a grid search using the best evaluation score on a held-out development set,
or by applying an approximate automatic tuning procedure based on a heuristic guess
of which direction the $\alpha$ parameter should be adjusted.

\subsection{Greedy Unigram Likelihood} % Segmentation}

\newcite{varjokallio2013learning} presents a subword segmentation method,
particularly designed for use in ASR.
It applies greedy pruning based on unigram likelihood.
The seed lexicon is constructed by enumerating all substrings from a list of common words,
up to a specified maximum length.
Pruning proceeds in two phases,
which the authors call \emph{initialization} and \emph{pruning}.

In the first phase,
a character-level language model is trained.
The initial probabilities of the subwords are computed using the language model.
The probabilities are refined by EM, followed by hard-EM.
During the hard-EM, frequency based pruning of subwords begins.

In the second phase,
hard-EM is used for parameter estimation.
At the end of each iteration,
the least frequent subwords are selected as candidates for pruning.
For each candidate subword,
the change in likelihood when removing the subword
is estimated by resegmenting all words in which the subword occurs.
After each pruned subword, the parameters of the model are updated.
Pruning ends when the goal lexicon size is reached or
the change in likelihood no longer exceeds a given threshold.

%% already summarized in the table
%In short, this method is a maximum likelihood EM+Prune
%with a heuristic iteration structure,
%and a heuristic stopping condition based on thresholding the change in likelihood.

%%%% floats
% takes too much space
%\input{table_train_variants.tex}

% not sure about these (some, none or all?)
%\figcost{eng}{English}{1.0}
%\figcost{fin}{Finnish}{1.0}
%\figcost{tur}{Turkish}{0.91}
%\figcost{sme}{North Sámi}{0.91}
\costsubfigs{}
%%%% floats

\subsection{SentencePiece}

SentencePiece~\cite{kudo2018sentencepiece,kudo2018subword}
is a subword segmentation method aimed for use in any NLP system,
particularly NMT.
One of its design goals is use in multilingual systems.

% Should this be in the EM section?
Although~\cite{kudo2018subword} implies a use of maximum likelihood estimation,
the reference implementation\footnote{https://github.com/google/sentencepiece} uses the implicit Dirichlet Process prior
called Bayesian EM~\cite{liang2007structured}.  % "How to Bayesianify your EM"
% sparsity inducing
In the M-step, the count normalization is modified to
\begin{equation}
\prob(z) = \frac{ \exp(\Psi(C_{z})) }{ \exp(\Psi( \sum_{z'} C_{z'} )) }
\end{equation}
where $\Psi$ is the digamma function.

The seed lexicon is simply the e.g. one million most frequent substrings.
% seems to be partial redundancy removal, but paper says nothing about it
%
SentencePiece uses an EM+Prune training algorithm.
Each iteration consists of two sub-iterations of EM,
after which the lexicon is pruned.
Pruning is based on Viterbi counts (EM+Viterbi-prune).
First, subwords that do not occur in the Viterbi segmentation are pre-pruned.
The cost function is the estimated change in likelihood when the subword is removed,
estimated using the assumption that all probability mass of the removed subword goes to its Viterbi segmentation.
Subwords are sorted according to the cost,
and a fixed proportion of remaining subwords are pruned each iteration.
Single character subwords are never pruned.
A predetermined lexicon size is used as the stopping condition.

\section{Morfessor EM+Prune}

Morfessor EM+Prune\footnote{Software available at \url{https://github.com/Waino/morfessor-emprune} .}
uses the unigram language model and priors similar to Morfessor
Baseline, but combines them with EM+Prune training.

\subsection{Prior}
The prior must be slightly modified for use with the EM+Prune algorithm.
The prior for the frequency distribution \eqref{eq:freqdistr} is derived using combinatorics.
When using real-valued expected counts,
there are infinite assignments of counts to parameters.
Despite not being theoretically motivated,
it can still be desirable to compute an approximation of the Baseline frequency distribution prior,
in order to use EM+Prune as an improved search to find more optimal parameters for the original cost.
To do this, the real valued token count $\nu$ is rounded to the nearest integer%
\footnote{An alternative would be to replace the factorial with the gamma function.
This added precision serves no practical purpose,
particularly as we already use Stirling's approximation of the factorial.}.
Alternatively, the prior for the frequency distribution can be omitted,
or a new prior with suitable properties could be formulated.
We do not propose a completely new prior in this work,
instead opting to remain as close as possible to Morfessor Baseline.

In Morfessor EM+Prune,
morphs are explicitly stored in the lexicon,
and morphs are removed from the lexicon only during pruning.
This differs from Morfessor Baseline,
in which a morph is implicitly considered to be stored in the lexicon if it has non-zero count.

The prior for the morph form properties does not need to be modified.
% mention permutation cost?
During the EM parameter estimation, the prior for the morph form properties is omitted
as the morph lexicon remains constant.
During pruning, the standard form prior is applicable.

Additionally we apply the Bayesian EM implicit Dirichlet Process prior~\cite{liang2007structured}.
We experiment with four variations of the prior:
\begin{enumerate}
    \item the full EM+Prune prior, % bl
    %\item the EM+Prune prior without Bayesian EM (noexp$\Psi$), % noexpdigamma
    \item omitting the Bayesian EM (noexp$\Psi$), % noexpdigamma
    \item omitting the approximate frequency distribution prior (nofreqdistr), % nofreqdistr
    \item and omitting the prior entirely (noprior). % nolexcost
\end{enumerate}
%Omitting the prior: as if tuning $\alpha \to \infty$, but in practice the formula is modified and no weighting is applied.

\subsection{Seed Lexicon}
The seed lexicon consists of the one million most frequent substrings,
with two restrictions on which substrings to include: pre-pruning of redundant subwords, and forcesplit.
Truncating to the chosen size is performed after pre-pruning,
which means that pre-pruning can make space for substrings that would otherwise have been below the threshold.

Pre-pruning of \term{redundant subwords} is based on occurrence counts.
If a string $x$ occurs $n$ times, then any substring of $x$ will occur at least $n$ times.
Therefore, if the substring has a count of exactly $n$,
we know that it is not needed in any other context except as a part of $x$.
Such unproductive substrings are likely to be poor candidate subwords,
and can be removed to make space in the seed lexicon for more useful substrings.
This pre-pruning is not a neutral optimization, but does affect segmentation results.
% Might be that the longer string is e.g. stem+suffix, where the stem is not seen with any other suffix.
% The stem+suffix would still likely be selected for the lexicon as it can be emitted as a single unit.
% The stem would only be recovered if the suffix is common enough that its corpus cost is less than the extra lexicon cost of storing the longer unit.
We check all initial and final substrings for redundancy, but do not pre-prune internal substrings.

To achieve \term{forced splitting} before or after certain characters,
e.g. hyphens, apostrophes and colons,
substrings which include a forced split point can be removed from the seed lexicon.
As EM+Prune is unable to introduce new subwords, this pre-pruning is sufficient to guarantee the forced splits.
While Morfessor 2.0 only implements force splitting certain characters to single-character morphs, i.e. force splitting on both sides,
we implement more fine-grained force splitting separately before and after the specified character.

%%%% floats
% include these, but culled, maybe merged
\begin{table}[t]
\centering
%\resizebox{\columnwidth}{!}{
\scalebox{0.73}{
\begin{tabular}{lrcrrr}
\toprule
                                                & FS          &  Prior \dda  &  Likelihood \dda  & W-sum \dda \\
 \midrule
 Words                                          &             &     1.79\sci7  &    1.32\sci7  &    2.98\sci7  \\
 Characters                                     &             &     2.35\sci2  &    2.90\sci7  &    2.61\sci7  \\
 \midrule
 \bb EM+P     MDL                               & \checkmark  & \bb 4.69\sci5  &    2.09\sci7  &    1.92\sci7  \\
 Morfessor Baseline                             & \checkmark  &     7.55\sci5  &    2.05\sci7  &    1.92\sci7  \\
 Morfessor Baseline                             &             &     8.84\sci5  &\bb 1.99\sci7  &\bb 1.88\sci7  \\
 EM+P     MDL                                   &             &     5.80\sci5  &    2.02\sci7  &\bb 1.88\sci7  \\
 EM+P     MDL lateen                            &             &     6.35\sci5  &    2.01\sci7  &\bb 1.88\sci7  \\
\bottomrule
\end{tabular}}
\caption{Morfessor cost results for English. $\alpha=0.9$.
FS is short for forcesplit, W-sum for weighted sum of prior and likelihood.
\dda means that lower values are better.
The bolded method is our primary configuration.
\label{tab:cost:eng}}
\end{table}

\begin{table}[t]
\centering
%\resizebox{\columnwidth}{!}{
\scalebox{0.73}{
\begin{tabular}{lrcrrr}
\toprule
                                                & FS          &  Prior \dda  &  Likelihood \dda  & W-sum \dda \\
 \midrule
 Words                                            &             &    8.64\sci7  &    4.77\sci7  &     8.74\sci7  \\
 Characters                                       &             &    2.46\sci2  &    1.27\sci8  &     2.54\sci6  \\
 \midrule
 Morfessor Baseline                               & \checkmark  &\bb 8.31\sci4  &    8.60\sci7  &     1.80\sci6  \\
 Morfessor Baseline                               &             &    8.36\sci4  &    8.59\sci7  &     1.80\sci6  \\
 \bb EM+P     MDL                                 & \checkmark  &    1.29\sci5  &    8.28\sci7  &     1.79\sci6  \\
 EM+P     MDL lateen                              &             &    1.41\sci5  &\bb 8.22\sci7  &     1.79\sci6  \\
 EM+P     MDL                                     &             &    1.31\sci5  &    8.26\sci7  &\bb  1.78\sci6  \\
\bottomrule
\end{tabular}}
\caption{Morfessor cost results for Finnish. $\alpha=0.02$.
%FS is short for forcesplit.
\label{tab:cost:fin}}
\end{table}

\begin{table}[t]
\centering
%\resizebox{\columnwidth}{!}{
\scalebox{0.73}{
\begin{tabular}{lrcrrr}
\toprule
                                                &  Prior \dda  &  Likelihood \dda  & W-sum \dda \\
 \midrule
 Words                                          &     1.31\sci7  &    9.09\sci6  &    1.68\sci7  \\
 Characters                                     &     1.19\sci2  &    2.08\sci7  &    8.30\sci6  \\
 \midrule
 Morfessor Baseline                             & \bb 2.54\sci5  &    1.39\sci7  &    5.82\sci6  \\
 EM+P     MDL lateen                            &     2.79\sci5  &    1.37\sci7  &    5.78\sci6  \\
 \bb EM+P     MDL                               &     2.71\sci5  &    1.37\sci7  &    5.77\sci6  \\
 EM+P     MDL      keep-redundant               &     2.97\sci5  &\bb 1.36\sci7  &\bb 5.73\sci6  \\
\bottomrule
\end{tabular}}
\caption{Morfessor cost results for Turkish. $\alpha=0.4$
\label{tab:cost:tur}}
\end{table}

\begin{table}[t]
\centering
%\resizebox{\columnwidth}{!}{
\scalebox{0.73}{
\begin{tabular}{lrcrrr}
\toprule
                                                & FS          &  Prior \dda  &  Likelihood \dda  & W-sum \dda \\
 \midrule
 Words                                          &             &     2.12\sci7  &    1.03\sci7  &    3.15\sci7  \\
 Characters                                     &             &     1.38\sci3  &    2.98\sci7  &    2.98\sci7  \\
 \midrule
 Morfessor Baseline                             & \checkmark  &     1.76\sci6  &    1.62\sci7  &    1.80\sci7  \\
 Morfessor Baseline                             &             &     1.87\sci6  &\bb 1.61\sci7  &    1.80\sci7  \\
 \bb EM+P     MDL                               & \checkmark  &     9.52\sci5  &    1.70\sci7  &\bb 1.79\sci7  \\
 EM+P     MDL lateen                            &             &     9.83\sci5  &    1.69\sci7  &\bb 1.79\sci7  \\
 EM+P     MDL                                   &             &     9.56\sci5  &    1.69\sci7  &\bb 1.79\sci7  \\
\bottomrule
\end{tabular}}
\caption{Morfessor cost results for North Sámi. $\alpha=1.0$
%FS is short for forcesplit.
\label{tab:cost:sme}}
\end{table}

%%%% floats

\subsection{Training Algorithm}

We experiment with three variants of the EM+Prune iteration structure:
\begin{enumerate}
    \item EM,
    \item Lateen-EM,
    \item EM+Viterbi-prune
\end{enumerate}

EM+Viterbi-prune is
an intermediary mode between EM and lateen-EM in the context of pruning.
The pruning decisions are made based on counts from a single iteration of Viterbi training,
but these Viterbi counts are not otherwise used to update the parameters.
%When entering the pruning phase, a copy of the soft EM parameters is stored.
%The pruning decisions are made based on counts from a 
%single iteration of Viterbi training.
%For the subwords that were not pruned, the counts are reset to the stored soft EM parameters.
In effect, this allows for the more aggressive pruning using the Viterbi counts,
while retaining the uncertainty of the soft parameters.

Each iteration begins with 3 sub-iterations of EM.
In the pruning phase of each iteration,
the subwords in the current lexicon are sorted in ascending order according to the estimated change in the cost function if the subword is removed from the lexicon.
Subwords consisting of a single character are always kept, to retain the ability to represent an open vocabulary without OOV issues.
The list is then pruned according to one of three available pruning criteria:%
\footnote{MDL with or without automatic tuning is not compatible with omitting the prior.}
\begin{enumerate}
    \item ($\alpha$-weighted) MDL pruning,
    \item MDL with automatic tuning of $\alpha$ for lexicon size,
    \item lexicon size with omitted prior or pretuned $\alpha$.
\end{enumerate}
%Table \ref{tab:trainvariants} shows the compatibility of prior settings with pruning criteria.

In ($\alpha$-weighted) Minimum Description Length (MDL) pruning,
subwords are pruned until the estimated cost starts rising, or until the pruning quota for the iteration is reached, whichever comes first.

A subword lexicon of a predetermined size can be used as pruning criterion in two different ways.
If the desired $\alpha$ is known in advance, or if the prior is omitted,
subwords are pruned until the desired lexicon size is reached, or until the pruning quota for the iteration is reached, whichever comes first.

To reach a subword lexicon of a predetermined size while using the Morfessor prior,
the new \term{automatic tuning} procedure can be applied.
For each subword, the estimated change in prior and likelihood are computed separately.
These allow computing the value of $\alpha$ that would cause the removal of each subword to be cost neutral,
i.e. the value that would cause MDL pruning to terminate at that subword.
For subwords with the same sign for both the change in prior and likelihood,
no such threshold $\alpha$ can be computed:
if the removal decreases both costs the subword will always be removed,
and if it increases both costs it will always be kept.
Sorting the list of subwords according to the estimated threshold $\alpha$ including the always kept subwords allows
automatically tuning $\alpha$ so that a subword lexicon of exactly the desired size is retained after MDL pruning.
The automatic tuning is repeated before the pruning phase of each iteration, as retraining the parameters affects the estimates.

%%%% floats
% include these, but culled
\begin{table}[t]
\centering
\resizebox{\columnwidth}{!}{
\begin{tabular}{lrcrrrc@{}c}
\toprule
                                                     & $\alpha$ & FS         & Pre \uua  & Rec \uua&  F \uua &       &        \\ %& Type-acc    Pre   Rec     F   Acc
\midrule
 EM+P     MDL      noexp$\Psi$                       &     0.8  & \checkmark &    82.9  &    71.8 &\bb 77.0 &        &        \\ %&     0.52  0.85  0.73  0.79  0.51
 EM+P     MDL      nofreqdistr                       &     0.8  & \checkmark &    83.3  &    71.4 &    76.9 &        &        \\ %&     0.52  0.85  0.73  0.79  0.51
%EM+P     MDL      keep-redundant                    &     0.8  & \checkmark &    81.6  &    72.7 &    76.9 & \nonsE &        \\ %&     0.52  0.83  0.74  0.78  0.50
%EM+P     MDL      red5                              &     0.7  & \checkmark &    80.5  &\bb 73.4 &    76.8 & \nonsE &        \\ %&     0.52  0.81  0.74  0.78  0.49
\bb EM+P     MDL                                     &     0.9  & \checkmark &    81.9  &    72.1 &    76.7 & --     &        \\ %&     0.52  0.84  0.74  0.79  0.50
%EM+P     auto          38k                          &     1.0  & \checkmark &    81.6  &    72.2 &    76.6 & \nonsE &        \\ %&     0.52  0.84  0.73  0.78  0.51
\bb Morfessor Baseline                               &     0.8  & \checkmark &\bb 85.0  &    68.5 &    75.9 &        & --     \\ %&     0.51  0.86  0.69  0.77  0.48
 Morfessor Baseline                                  &     0.7  &            &    83.8  &    69.4 &    75.9 &        & \nonsB \\ %&     0.50  0.86  0.69  0.77  0.47
 EM+P     MDL                                        &     0.6  &            &    79.0  &\bb 72.8 &    75.8 &        &        \\ %&     0.49  0.81  0.72  0.77  0.45
%EM+P     MDL lateen                                 &     0.9  & \checkmark &    83.0  &    69.4 &    75.6 &        &        \\ %&     0.51  0.85  0.71  0.77  0.50
 SentencePiece 50k                                   &     --   &            &    75.9  &    61.9 &    68.2 &        &        \\ %&     0.34  0.75  0.63  0.68  0.32

\bottomrule
\end{tabular}}
\caption{Boundary Precision (Pre), Recall (Rec) and F$_{1}$-score (F) results for
English.
\nonsE\ indicates \textbf{not} significantly different (two-sided Wilcoxon signed-rank test, $p<0.05$, zero splitting)
from the bolded EM+Prune method,
and \nonsB\ from the bolded Baseline.
\label{tab:bpr:eng}}
\end{table}

\begin{table}[t]
\centering
\resizebox{\columnwidth}{!}{
\begin{tabular}{lrcrrrc@{}c}
\toprule
                                                      & $\alpha$ & FS         & Pre \uua  & Rec \uua&  F \uua  &       &         \\ %& Type-acc    Pre   Rec     F   Acc
 \midrule
\bb EM+P     MDL                                      &    0.035 & \checkmark &     72.0 &    55.8 & \bb 62.9 & --     &        \\ % &     0.21  0.77  0.60  0.68  0.24
 EM+P     MDL      nofreqdistr                        &     0.02 & \checkmark &     68.7 &    58.0 & \bb 62.9 & \nonsE &        \\ % &     0.21  0.75  0.63  0.68  0.24
%EM+P     MDL      keep-redundant                     &     0.01 & \checkmark &     64.3 &\bb 61.4 &     62.8 & \nonsE &        \\ % &     0.19  0.70  0.67  0.68  0.21
%EM+P     MDL      red5                               &      0.1 & \checkmark &     78.7 &    52.1 &     62.7 & \nonsE &        \\ % &     0.22  0.83  0.55  0.66  0.23
 EM+P     MDL      noexp$\Psi$                        &     0.02 & \checkmark &     68.4 &    57.9 &     62.7 & \nonsE &        \\ % &     0.21  0.74  0.63  0.68  0.24
 EM+P     MDL                                         &    0.015 &            &     66.7 &\bb 58.5 &     62.3 & \nonsE &        \\ % &     0.20  0.73  0.64  0.68  0.23
%EM+P     MDL lateen                                  &     0.02 & \checkmark &     68.9 &    56.4 &     62.0 &        & \nonsB \\ % &     0.20  0.75  0.62  0.68  0.23
\bb Morfessor Baseline                                &     0.02 & \checkmark &     62.3 &    58.2 &     60.2 &        & --     \\ % &     0.17  0.68  0.62  0.65  0.19
 SentencePiece 50k                                    &       -- &            & \bb 75.7 &    49.3 &     59.7 &        & \nonsB \\ % &     0.17  0.77  0.50  0.61  0.16
 Morfessor Baseline                                   &     0.02 &            &     62.0 &    57.6 &     59.7 &        &        \\ % &     0.17  0.68  0.62  0.65  0.20

\bottomrule
\end{tabular}}
\caption{Boundary Precision (Pre), Recall (Rec) and F$_{1}$-score (F) results for
 Finnish.
\label{tab:bpr:fin}}
\end{table}

\begin{table}[t]
\centering
\resizebox{\columnwidth}{!}{
\begin{tabular}{lrrrrc@{}c}
\toprule
                                                      & $\alpha$ & Pre \uua  & Rec \uua&  F \uua  &      &          \\ %& Type-acc    Pre   Rec     F   Acc
 \midrule
 EM+P     MDL      keep-redundant                     &     0.3  & \bb 87.8 &    58.7 &\bb 70.4 &        &        \\ % &     0.27  0.92  0.56  0.70  0.24
%EM+P     MDL      red5                               &     0.2  &     85.6 &    59.3 &    70.1 &        &        \\ % &     0.26  0.89  0.55  0.68  0.23
 EM+P     MDL      noexp$\Psi$                        &     0.4  &     87.6 &    58.1 &    69.9 &        &        \\ % &     0.27  0.90  0.55  0.68  0.24
 EM+P     MDL      nofreqdistr                        &     0.3  &     86.4 &    58.2 &    69.6 & \nonsE &        \\ % &     0.27  0.90  0.55  0.68  0.23
%EM+P     auto          15k                           &       ?  &     84.8 &    58.7 &    69.4 & \nonsE &        \\ % &     0.25  0.89  0.56  0.69  0.23
\bb EM+P     MDL                                      &     0.2  &     84.8 &    58.7 &    69.4 & --     &        \\ % &     0.25  0.89  0.56  0.69  0.23
%EM+P     MDL lateen                                  &     0.2  &     84.8 &    58.6 &    69.3 & \nonsE &        \\ % &     0.25  0.89  0.55  0.68  0.22
\bb Morfessor Baseline                                &     0.2  &     78.2 &    58.4 &    66.9 &        & --     \\ % &     0.23  0.82  0.56  0.66  0.22
 SentencePiece 12k                                    &      --  &     75.2 &\bb 60.0 &    66.8 &        & \nonsB \\ % &     0.21  0.80  0.57  0.67  0.21

\bottomrule
\end{tabular}}
\caption{Boundary Precision (Pre), Recall (Rec) and F$_{1}$-score (F) results for
Turkish.
\label{tab:bpr:tur}}
\end{table}

\begin{table}[t]
\centering
\resizebox{\columnwidth}{!}{
\begin{tabular}{lrcrrrc@{}c}
\toprule
                                                     & $\alpha$ & FS         & Pre \uua  & Rec \uua&  F \uua  &       &         \\ %& Type-acc    Pre   Rec     F   Acc
 \midrule
\bb Morfessor Baseline                               &     1.4  &            &\bb  75.7 &    60.7 &\bb  67.4 & \nonsE & --     \\ %&     0.40  0.77  0.68  0.72  0.49
 EM+P     MDL      nofreqdistr                       &     1.0  & \checkmark &     73.7 &    61.8 &     67.2 &        & \nonsB \\ %&     0.42  0.75  0.70  0.73  0.52
 Morfessor Baseline                                  &     1.2  & \checkmark &     75.7 &    60.4 &     67.2 & \nonsE & \nonsB \\ %&     0.41  0.77  0.68  0.72  0.50
\bb EM+P     MDL                                     &     1.3  & \checkmark &     73.0 &    62.1 &     67.1 & --     & \nonsB \\ %&     0.42  0.75  0.70  0.73  0.52
%EM+P     auto          75k                          &     1.0  & \checkmark &     73.1 &    61.9 &     67.0 & \nonsE &        \\ %&     0.42  0.74  0.69  0.72  0.50
 EM+P     MDL                                        &     1.2  &            &     72.8 &    62.0 &     66.9 &        &        \\ %&     0.42  0.74  0.70  0.72  0.50
 EM+P     MDL      noexp$\Psi$                       &     0.4  & \checkmark &     66.5 &\bb 65.9 &     66.2 &        &        \\ %&     0.41  0.68  0.70  0.69  0.47
%EM+P     MDL      keep-redundant                    &     1.2  & \checkmark &     71.9 &    60.0 &     65.4 &        &        \\ %&     0.41  0.74  0.69  0.72  0.52
%EM+P     MDL lateen                                 &     1.3  & \checkmark &     73.0 &    59.2 &     65.4 &        &        \\ %&     0.40  0.74  0.68  0.71  0.50
%EM+P     MDL      red5                              &     0.4  & \checkmark &     63.9 &\bb 66.1 &     65.0 &        &        \\ %&     0.41  0.71  0.72  0.72  0.51
 SentencePiece 64k                                   &      --  &            &     65.3 &    61.3 &     63.3 &        &        \\ %&     0.40  0.71  0.69  0.70  0.50

\bottomrule
\end{tabular}}
\caption{Boundary Precision (Pre), Recall (Rec) and F$_{1}$-score (F) results for
 North Sámi.
\label{tab:bpr:sme}}
\end{table}

% perhaps not include these: figbpr shows more
%\figfscore{eng}{English}
%\figfscore{fin}{Finnish}
%\figfscore{tur}{Turkish}
%\figfscore{sme}{North Sámi}

% include these (some or all?)
\bprsubfigs{}
%\figbpr{eng}{English}{The smallest and largest $\alpha$-values are labeled.}
%\figbpr{fin}{Finnish}{The smallest and largest $\alpha$-values are labeled.}
%\figbpr{tur}{Turkish}{}
%\figbpr{sme}{North Sámi}{}
%%%% floats

\subsection{Sampling of Segmentations}

Morfessor EM+Prune can be used in subword regularization~\cite{kudo2018subword},
a denoising-based regularization method for neural NLP systems.
Alternative segmentations can be sampled
from the full data distribution using Forward-filtering backward-sampling algorithm~\cite{scott2002bayesian}
or approximatively but more efficiently from an $n$-best list.

\subsection{SentencePiece as a Special Case of Morfessor EM+Prune}

Table \ref{tab:comparison} contains a comparison between all four methods discussed in this work.
To recover SentencePiece, Morfessor EM+Prune should be run with the following settings:
The prior should be omitted entirely, leaving only the likelihood
\begin{equation}
%\hparams = \argmin_{\params} \cost(\params, \data) = \argmin_{\params} - \log\prob(\params) - \alpha\log\prob(\data \vb \params)
\hparams = \argmin_{\params} \{
    - \log\prob(\data \vb \params) \}
\end{equation}
As the tuning parameter $\alpha$ is no longer needed when the prior is omitted,
the pruning criterion can be set to a predetermined lexicon size, without automatic tuning of $\alpha$.
Morfessor by default uses type-based training;
to use frequency information, count dampening should be turned off.
The seed lexicon should be constructed without using forced splitting.
The EM+Viterbi-prune training scheme should be used,
with Bayesian EM turned on.

\section{Experimental Setup}

English, Finnish and Turkish data are from the \emph{Morpho Challenge 2010} data set~\cite{kurimo2010morpho,kurimo2010overview}.
%\citelanguageresource{morphochal10}.
The training sets contain ca 878k, 2.9M and 617k word types, respectively.
As test sets we use the union of the 10 official test set samples.
For North Sámi, we use a list of ca 691k word types extracted from \emph{Den samiske tekstbanken} corpus~(Sametinget, 2004)
%\footnote{Provided by UiT, The Arctic University of Norway.},       % LREC has specific format for language resources
and the 796 word type test set from version 2 of the data set collected by~\cite{gronroos2015low-resource,gronroos2016low-resource}.
%The North Sámi unannotated training data we use does not contain word counts.   % so also SentencePiece is trained on types

In most experiments we use a grid search with a development set to find a suitable value for $\alpha$.
The exception is experiments using autotuning or lexicon size criterion, and experiments using SentencePiece.
We use type-based training (dampening counts to 1) with all Morfessor methods.

For English, we force splits before and after hyphens, and before apostrophes,
e.g. \examp{women's-rights} is force split into \examp{women \tbnd 's \tbnd - \tbnd rights}.
For Finnish, we force splits before and after hyphens, and after colons.
For North Sámi, we force splits before and after colons.
For Turkish, the Morpho Challenge data is preprocessed in a way that makes force splitting ineffectual.

\subsection{Evaluation}

The ability of the training algorithm to find parameters minimizing the Morfessor cost
is evaluated by using the trained model to segment the training data,
and loading the resulting segmentation as if it was a Morfessor Baseline model.
We observe both unweighted prior and likelihood,
and their $\alpha$-weighted sum.

The closeness to linguistic segmentation is evaluated by comparison
with annotated morph boundaries using \emph{boundary precision}, \emph{boundary recall},
and \emph{boundary $F_{1}$-score}~\cite{virpioja2011empirical}.
The boundary $F_{1}$-score (F-score for short) equals the harmonic mean of precision
(the percentage of correctly assigned boundaries with respect to all assigned boundaries)
and recall (the percentage of correctly assigned boundaries with respect to the reference boundaries).
%
%\begin{align}
%\begin{split}
%\mathrm{Precision} = \frac{\occ(\mathrm{correct})}{\occ(\mathrm{proposed})};
%\end{split}
%\begin{split}
%\mathrm{Recall} = \frac{\occ(\mathrm{correct})}{\occ(\mathrm{reference})}
%\label{eq:bpr}
%\end{split}
%\end{align}
%where $\occ$ is the occurrence count.
%
Precision and recall are calculated using macro-averages over the word types in the test set.
In the case that a word has more than one annotated segmentation, we take the one that gives the highest score.

% Perhaps drop type level accuracy from the result tables? No reason to compare to iwclul 2019 paper, which was semi-sup.
% A word in the test set is counted as correct if all boundary decisions are correct.

\subsection{Error Analysis} % or describe it along with the Results?

We perform an error analysis, with the purpose of gaining more insight into the
ability of the methods to model particular aspects of morphology.
We follow the procedure used by~\newcite{ruokolainen2016comparative}.
It is based on a categorization of morphs into the categories \techterm{prefix}, \techterm{stem}, and \techterm{suffix}.
The category labels are derived from the original morphological analysis labels in the English and Finnish gold standards,
and directly correspond to the annotation scheme used in the North Sámi test set.

We first divide errors into two kinds, \emph{over-segmentation} and \emph{under-segmentation}.
Over-segmentation occurs when a boundary is incorrectly assigned within a morph segment.
In under-segmentation, the a correct morph boundary is omitted from the generated segmentation.
We further divide the errors by the morph category in which the over-segmentation occurs,
and the two morph categories surrounding the omitted boundary in under-segmentation.

\section{Results}

%%%%%%%%%% Old preliminary results
%% eng: sometimes better F,    always better cost, not top
%% fin:   usually better F, sometimes better cost, coincide
%% tur:    always better F,   usually better cost, coincide
%% sme: sometimes better F,   usually better cost, not top
% There is a range of tuning for which EM+Prune finds better cost.
% For Finnish and Turkish, higher F is reached.
% 
% EM+Prune is not as responsive to extreme tuning as Morfessor Baseline,
% in particular very high values for alpha.
% EM+Prune is not able to add subwords to increase likelihood, if they are not present in the seed lexicon.
%%%%%%%%%% Old preliminary results end

%\subsection{Cost optimization}

Figure~\ref{fig:costs} compares the cost components of the Morfessor
model across different $\alpha$ parameters. The lowest costs for the
mid-range settings are obtained for the EM+Prune algorithm, but for
larger lexicons, the Baseline algorithm copes better. As expected,
using forced splits at certain characters increase the costs, and the
increase is larger than between the training algorithms.
As Turkish preprocessing causes the results to be unaffected by the forced splits,
we only report results without them.

Tables~\ref{tab:cost:eng} to \ref{tab:cost:sme}
show the Morfessor cost of the segmented training data for particular $\alpha$ values.
Again, the proposed Morfessor EM+Prune reaches a lower Morfessor cost than Morfessor Baseline. Using the lateen-EM has only minimal effect to the costs, decreasing the total cost slightly for English and increasing for the other languages.
Turkish results include the ``keep-redundant'' setting discussed below in more detail.

%\subsection{Linguistic evaluation}

Figure \ref{fig:bprs} shows the Precision--Recall curves for the primary systems,
for all four languages.
While increasing the Morfessor cost, 
forced splitting improves BPR.
Tables \ref{tab:bpr:eng} to \ref{tab:bpr:sme} show test set Boundary Precision, Recall and F$_{1}$-score (BPR) results
at the optimal tuning point (selected using a development set) for each model,
for English, Finnish, Turkish and North Sámi, respectively%.
\footnote{Note that SentencePiece is not designed for %unsupervised learning
aiming towards a linguistic morpheme segmentation.
Neither does it attempt to minimize the Morfessor cost.
Therefore, SentencePiece is included in the evaluations for context, not as a baseline method.}.
The default Morfessor EM+Prune configuration (``soft'' EM, full prior, forcesplit) significantly outperforms Morfessor Baseline w.r.t. the F-score
for all languages except North Sámi, for which there is no significant difference between the methods.
% North Sámi: tied with baseline

Morfessor EM+Prune is less responsive to tuning than Morfessor Baseline.
This is visible in the shorter lines in Figures~\ref{fig:costs} and \ref{fig:bprs},
although the tuning parameter takes values from the same range.
In particular, EM+Prune can not easily be tuned to produce very large lexicons.

%We do not see an improvement in BPR from using lateen-EM or EM+Viterbi-prune. 

%\subsection{Seed lexicon}

% keep-redundant better for English (tiny difference) and Turkish, removing redundants better for Finnish and North Sámi.
Pre-pruning of redundant substrings gives mixed results.
For Turkish, both Morfessor cost and BPR are degraded by the pre-pruning,
but for the other three languages the pre-pruning is beneficial or neutral.
When tuning $\alpha$ to very high values (less segmentation),
pre-pruning of redundant substrings improves the sensitivity to tuning.
The same effect may also be achievable by using a larger seed lexicon.
We perform most of our experiments with pre-pruning turned on.

To see the effect of pre-pruning on the seed lexicon,
we count the number of subwords that are used in the gold standard segmentations, %for train and development sets
but not included in seed lexicons of various sizes.
Taking Finnish as an example,
we see 203 subword types missing from a 1 million substring seed lexicon without pre-pruning.
Turning on pre-pruning decreases the number of missing types to 120.
To reach the same number without using pre-pruning,
a much larger seed lexicon of 1.7M substrings must be used.

%\subsection{Prior components}

Omitting the frequency distribution appears to have little effect on Morfessor cost and BPR.
% English, Turkish and North Sámi see a tiny increase. Finnish is tied.
Turning off Bayesian EM (noexp$\Psi$) results in a less compact lexicon resulting in higher prior cost,
but improves BPR for two languages: English and Turkish.

%\subsection{Error analysis}

% eng: worse over, better under
% fin: better over, worse under (except SUF-SUF, UNKNOWN)
% sme: worse over, mixed under (better STM-SUF SUF-SUF, worse STM-STM, SUF-STM)
%Tables \ref{tab:error:fin} and \ref{tab:error:sme} contain the error analysis for Finnish and North Sámi.
Table \ref{tab:error} contains the error analysis for English, Finnish and North Sámi.
For English and North Sámi, EM+Prune results in less under-segmentation but worse over-segmentation.
For Finnish these results are reversed.
However, the suffixes are often better modeled,
as shown by lower under-segmentation on SUF-SUF (all languages) and STM-SUF (English and North Sámi).

% include these, but culled
%\input{table_analysis_fin.tex}
%\input{table_analysis_sme.tex}
\begin{table*}[t]
\centering
\resizebox{0.95\textwidth}{!}{
\begin{tabular}{llrrrrr|rrrrrrr}
\toprule
    &                    & \multicolumn{5}{c}{Over-segmentation}  & \multicolumn{7}{c}{Under-segmentation} \\
       \cmidrule(lr){3-7} \cmidrule(lr){8-14}
\vh{Language} &            Model   & \vh{STM} \dda  & \vh{SUF} \dda & \vh{PRE} \dda & \vh{UNKNOWN} \dda & \vh{\bf{PRE/TOT}} \uua & \vh{STM-SUF} \dda & \vh{STM-STM} \dda & \vh{SUF-SUF} \dda & \vh{SUF-STM} \dda & \vh{PRE-STM} \dda & \vh{UNKNOWN} \dda  & \vh{\bf{REC/TOT}} \uua \\
 
%                        & Over.STEM  & Over.SUFFIX  & Over.PREFIX  & Over.UNKNOWN  & Over.Total  & Under.STEM-SUFFIX  & Under.STEM-STEM  & Under.SUFFIX-SUFFIX  & Under.SUFFIX-STEM  & Under.PREFIX-STEM  & Under.UNKNOWN  & Under.Total
       \midrule
 eng& Characters         &     71.05  &       11.82  &        1.66  &         0.33  &      15.13  &              0.00  &            0.00  &                0.00  &              0.00  &              0.00  &                &      100.00  \\
 eng& Words              &      0.00  &        0.00  &        0.00  &         0.00  &     100.00  &             55.07  &            5.90  &                8.56  &              0.14  &              4.38  &                &       23.06  \\
       \midrule
 eng& SentencePiece 38k  &     17.60  &       10.25  &        0.18  &         0.24  &      71.74  &             26.40  &            2.48  &                2.74  &              0.05  &              2.78  &                &       65.26  \\
 eng& Morfessor Baseline & \bb 10.17  & \bb    2.32  & \bb    0.03  & \bb     0.07  & \bb  87.42  &             22.46  &            2.10  &                4.75  & \bb          0.04  &              1.65  &                &       67.37  \\
 eng& EM+Prune MDL       &     15.46  &        2.75  &        0.05  &         0.13  &      81.61  & \bb         19.93  & \bb        1.82  & \bb            4.32  & \bb          0.04  & \bb          1.46  &                & \bb   70.84  \\
       \midrule
    % eng: worse over, better under

%                        & Over.STEM  & Over.SUFFIX  & Over.PREFIX  & Over.UNKNOWN  & Over.Total  & Under.STEM-SUFFIX  & Under.STEM-STEM  & Under.SUFFIX-SUFFIX  & Under.SUFFIX-STEM  & Under.PREFIX-STEM  & Under.UNKNOWN  & Under.Total
       \midrule
 fin& Characters         &     65.23  &       13.80  &        0.67  &         0.57  &      19.73  &              0.00  &            0.00  &                0.00  &              0.00  &              0.00  &          0.00  &      100.00  \\
 fin& Words              &      0.00  &        0.00  &        0.00  &         0.00  &     100.00  &             49.19  &           17.16  &               21.76  &              4.84  &              0.96  &          0.58  &        4.09  \\
       \midrule
 fin& SentencePiece 13k  &     35.11  &        3.71  &        0.08  &         0.41  &      60.69  &             25.96  &            1.45  &               16.18  &              0.35  &              0.08  & \bb      0.16  & \bb   55.81  \\
 fin& Morfessor Baseline &     34.75  &        2.82  & \bb    0.03  &         0.38  &      62.02  & \bb         24.57  & \bb        0.86  &               16.31  & \bb          0.15  & \bb          0.04  &          0.20  &       57.63  \\
 fin& EM+Prune MDL       & \bb 29.34  & \bb    2.20  & \bb    0.03  & \bb     0.26  & \bb  68.18  &             24.68  &            0.90  & \bb           15.95  &              0.29  &              0.05  &          0.19  &       57.60  \\
       \midrule
     % fin: better over, worse under (except SUF-SUF, UNKNOWN)

%                        & Over.STEM  & Over.SUFFIX  & Over.PREFIX  & Over.UNKNOWN  & Over.Total &  Under.STEM-SUFFIX  & Under.STEM-STEM  & Under.SUFFIX-SUFFIX  & Under.SUFFIX-STEM  & Under.PREFIX-STEM &  Under.UNKNOWN  & Under.Total
       \midrule
 sme& Characters         &     81.44  &        6.80  &              &               &    11.76   &             0.00    &          0.00    &              0.00    &            0.00    &                   &                 &   100.00  \\
 sme& Words              &      0.00  &        0.00  &              &               &   100.00   &            52.92    &         13.15    &              4.43    &            0.61    &                   &                 &    28.64  \\
       \midrule
 sme& SentencePiece 64k  &     30.10  &        4.52  &              &               &    65.38   &            31.35    &          3.96    & \bb          3.09    &            0.20    &                   &                 &    61.40  \\
 sme& Morfessor Baseline & \bb 23.27  & \bb    3.02  &              &               &\bb 73.71   &            33.16    & \bb      2.22    &              3.40    & \bb        0.10    &                   &                 &    60.99  \\
 sme& EM+Prune MDL       &     23.35  &        4.41  &              &               &    72.25   & \bb        30.48    &          3.10    &              3.23    &            0.17    &                   &                 &\bb 62.84  \\
     % sme: worse over, mixed under (better STM-SUF SUF-SUF, worse STM-STM, SUF-STM)

\bottomrule
\end{tabular}}
\caption{Error analysis for
 English (eng, $\alpha=0.9$),
 Finnish (fin, $\alpha=0.02$), and
 North Sámi (sme, $\alpha=1.0$).
 All results without forcesplit.
 Over-segmentation and under-segmentation errors reduce precision and recall, respectively.
\label{tab:error}}
\end{table*}

\section{Conclusion}

We propose Morfessor EM+Prune,
a new training algorithm for Morfessor Baseline.
EM+Prune reduces search error during training, resulting in models with lower Morfessor costs.
Lower costs also lead to improved accuracy when segmentation output is compared to linguistic morphological segmentation.

We compare Morfessor EM+Prune to three previously published segmentation methods applying unigram language models.
We find that using the Morfessor prior is beneficial when the reference is linguistic morphological segmentation.

In this work we focused on model cost and linguistic segmentation.
In future work the performance of Morfessor EM+Prune in applications will be evaluated.
Also, a new frequency distribution prior, which is theoretically better motivated or has desirable properties, could be formulated.

\section{Acknowledgements}
% not doubleblind, should include ack directly

This study has been supported by the MeMAD project,
funded by the European Union's Horizon 2020
research and innovation programme~(grant agreement~\textnumero{}~780069),
and the FoTran project, funded by the European Research Council~(ERC) under the European Union's Horizon 2020 research and innovation programme~(grant agreement~\textnumero{}~771113)
Computer resources within the Aalto University School of Science ``Science-IT'' project were used.

%\footnotesize{
%This study has been supported by 
%two projects funded by the European Union's Horizon 2020 (H2020) research and innovation programme:
%the MeMAD project ~(grant~\textnumero{}~780069),
%and the FoTran project, funded by the European Research Council~(ERC) under the H2020 programme~(grant~\textnumero{}~771113).
%Computer resources within the Aalto University School of Science ``Science-IT'' project were used.}

% ugly hack: get the language resources to appear without replacing the cites with \citelanguageresource
%\phantom{
%\citelanguageresource{morphochal10}
%\citelanguageresource{tekstbanken}
%\citelanguageresource{smeactive}
%\citelanguageresource{morfessor2}
%\citelanguageresource{sentencepiece}}

% \nocite{*}
\section{Bibliographical References}\label{reference}

\bibliographystyle{lrec}
\bibliography{2020_lrec}

\section{Language Resource References}
%\label{lr:ref}
%\bibliographystylelanguageresource{lrec}
%\bibliographylanguageresource{languageresource}
{\leftskip=1em
\parindent=-1em
Grönroos, Stig-Arne and Hiovain, Katri and Smit, Peter and Rauhala,
Ilona Erika and Jokinen, Päivi Kristiina and Kurimo, Mikko and Virpioja, Sami. (2015).
\textit{North Sámi active learning morphological segmentation annotations}. Aalto University, 2.0.

Kudo, Taku and Richardson, John. (2018). \textit{SentencePiece}. Taku Kudo.

Kurimo, Mikko and Virpioja, Sami and Turunen, Ville T. (2010). \textit{Morpho Challenge 2010 dataset}. Aalto University.

Sametinget. (2004). \textit{Den samiske tekstbanken}. UiT Norgga árktalaš universitehta.

Virpioja, Sami and Smit, Peter and Grönroos, Stig-Arne and Kurimo, Mikko. (2013). \textit{Morfessor 2.0}. Aalto University, 2.0.6.

}

\end{document}